\documentclass[conference]{IEEEtran}

\IEEEoverridecommandlockouts
\usepackage{cite}
\usepackage{amsmath,amssymb,amsfonts}
\usepackage{algorithmic}
\usepackage{graphicx}
\usepackage[margin=0.7in]{geometry}
\usepackage{textcomp}
\usepackage[caption=false]{subfig}
\usepackage{float}
\usepackage{todonotes}
\def\BibTeX{{\rm B\kern-.05em{\sc i\kern-.025em b}\kern-.08em
    T\kern-.1667em\lower.7ex\hbox{E}\kern-.125emX}}
\begin{document}

\title{Surface Type Classification for Autonomous Robot Indoor Navigation
}
\author{\IEEEauthorblockN{Francesco Lomio$^*$\thanks{$^*$Corresponding author; email: francesco.lomio@tuni.fi}, Erjon Skenderi, Damoon Mohamadi, Jussi Collin, Reza Ghabcheloo and Heikki Huttunen}
\IEEEauthorblockA{
Tampere University, Finland}
}

\maketitle

\begin{abstract}
In this work we describe the preparation of a time series dataset of inertial measurements for determining the surface type under a wheeled robot. The data consists of over 7600 labeled time series samples, with the corresponding surface type annotation.
This data was used in two public competitions with over 1500 participant in total. 
Additionally, we describe the performance of state-of-art deep learning models for time series classification, as well as propose a baseline model based on an ensemble of machine learning methods. The baseline achieves an accuracy of over 68\% with our nine-category dataset.
\end{abstract}






\section{Introduction}
\label{sec:introduction}

Indoor and outdoor navigation has become an important concept for autonomous machines such as mobile robots and intelligent work machines. Most navigation techniques rely on external sources, such as geolocation based on Global Positioning System (GPS), Global Navigation Satellite System (GNSS), Bluetooth Low Energy (BLE) or WIFI. However, the navigation accuracy could be improved from novel data sources, that are not directly connected with the location, but serve as a proxy target to be used for data fusion. 

One such proxy measurement of location is the surface type underneath the autonomous vehicle. For example, the map may contain data of all surface types, and with an inertial measurement unit (IMU), once the measurement system detects a transition in the surface type, this fixes the location in the map, making the navigation data more reliable.

\begin{figure}[t!]
\centerline{\includegraphics[scale=0.40]{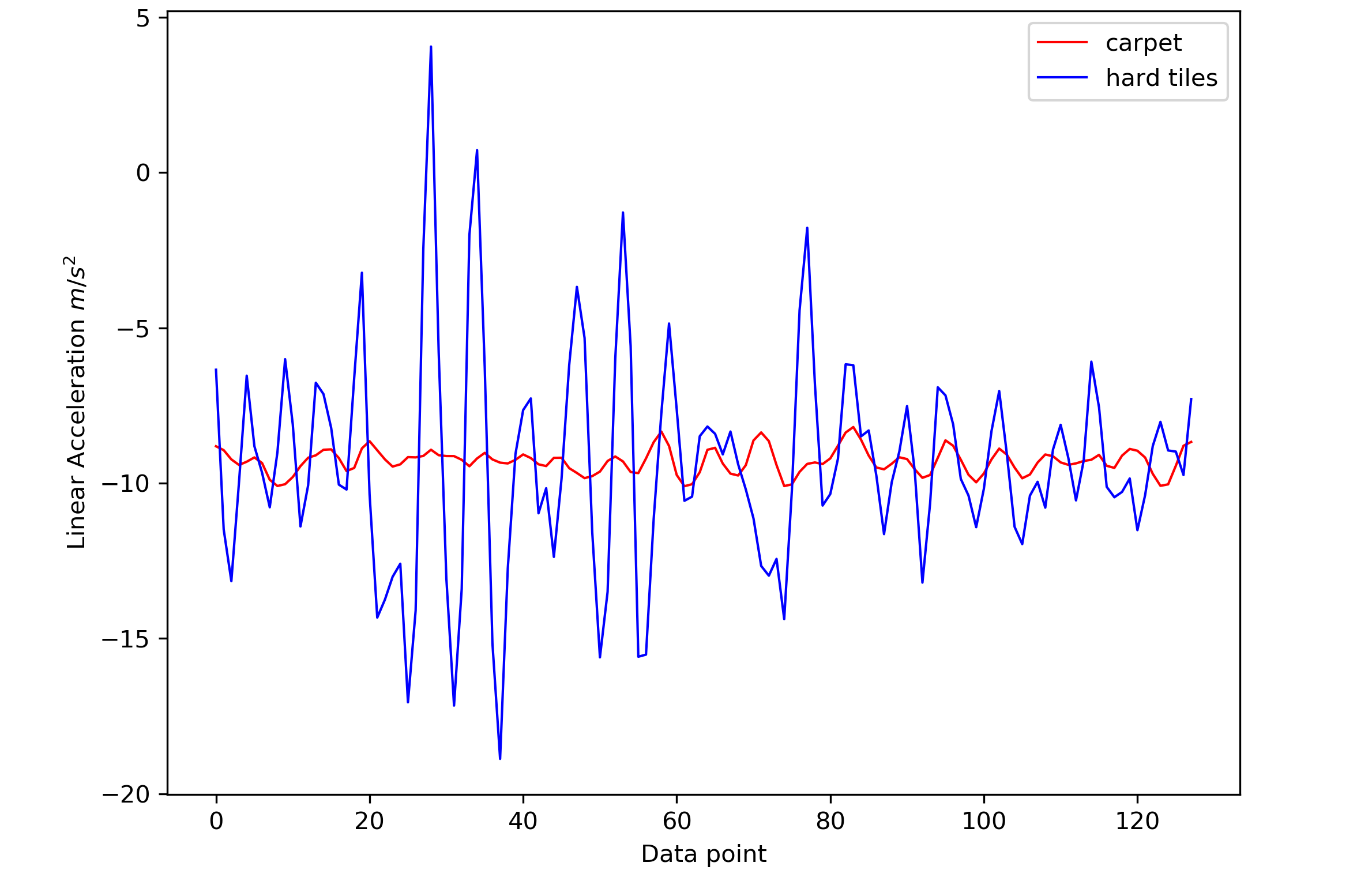}}
\caption{Variation of one of the measures recorded by the IMU for two classes.}
\label{fig_1}
\end{figure}

To this aim, this work describes a new dataset generated using an inertial measurement unit, whose variation can be used for surface type classification (Figure~\ref{fig_1}), along with a machine learning based approach for surface type classification.

Other work have shown the possibility of using this type of sensors for the surface recognition, using data related to outdoor surfaces~\cite{ojeda2006terrain}. In ~\cite{oliveira2017speed}, the authors used the Z-axis of the accelerometer of an IMU unit as input to classify the surface type for different velocity, but this was done only for outdoor data. In~\cite{kertesz2016rigidity}, the authors used sensors from legged walking robots to classify indoor surface. Other studies used a combination of multiple sensors and cameras~\cite{walas2015terrain}. In~\cite{brooks2012self}, the authors studied a method for a robot to learn to classify outdoor surface from vibration and traction sensors and use these information to learn to recognize the type of terrain ahead from an inbuilt camera.
Other approaches for the surface recognition include the use of audio data: in ~\cite{valada2017deep} the authors used the audio feedback from the interaction of the robot with the terrain (both indoor and outdoor) for classifying the surface type. 

We describe a procedure of data collection from nine indoor surface types collected using only an IMU sensor positioned on an industrial trolley with silent wheels. Each measurement sequence is linked with the actual surface. To the best of our knowledge, the dataset is unique both in its extent, as well as its scope. The data was used in two public machine learning competitions: first as a mandatory part of a university advanced machine learning course\footnote{\url{ www.kaggle.com/c/robotsurface}}; and secondly as a public competition for all participants of a worldwide \textit{CareerCon}\footnote{\url{ www.kaggle.com/c/career-con-2019}} event organized by Kaggle---a company whose platform is used for organizing large scale machine learning competitions. In total, over 1,500 people have participated in those competitions. 

\begin{figure*}[t]
\centerline{\includegraphics[width=1.0\textwidth]{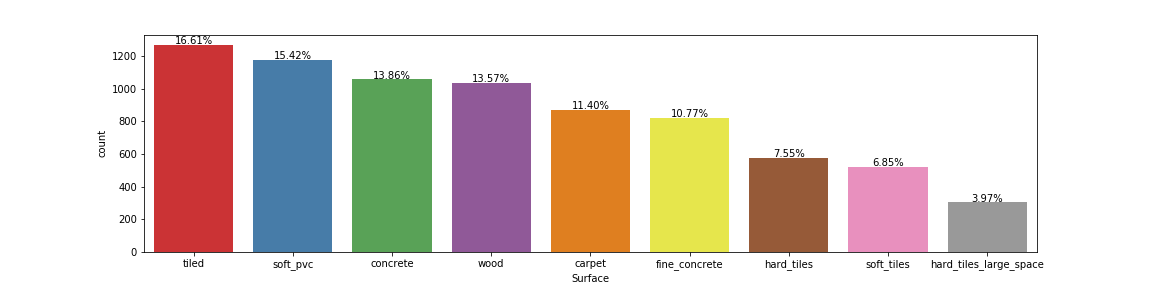}}
\caption{Sample distribution by class.}
\label{fig_2}
\end{figure*}

Beside introducing the dataset, in this work we also illustrate the results obtained by two state-of-art deep learning models for time series classification, tested on our proposed dataset, as well as a baseline model based on an ensemble of classical machine learning tool and deep learning.

Thus, to summarize, the main contributions of the paper are the following: (1) we release the most comprehensive dataset for indoor surface categorization o date and (2) we propose a baseline, which is fused from recent time series classification models and community methods that have proven successful in two competitions using our data. Finally, (3) the dataset is made publicly available in its entirety, which we hope will spread to public use within the machine learning community as a standard benchmark dataset\footnote{\url{ www.zenodo.org/record/2653918\#.XMgP2MRS-Uk}}.


In the remainder of this paper, we describe the data collection procedure in Section~\ref{sec:data} together with the characteristics of the data and a discussion of the preparation to a suitable form for a large scale competition. Next, Section~\ref{sec:methods} discusses the method used to test the data, specifically some state-of-art methods for time series classification. We also describe a baseline model, consisting of an ensemble of machine learning method and deep learning for the surface type classification. The results of the models used are discussed in Section~\ref{sec:evaluation}. Finally, Section~\ref{sec:conclusion} summarizes the results and discusses further work on this domain.

\section{Data collection and preparation}
\label{sec:data}
\subsection{Data collection}
\label{subsec:datacollection}

\begin{figure}[H]
\centerline{\includegraphics[scale=0.30]{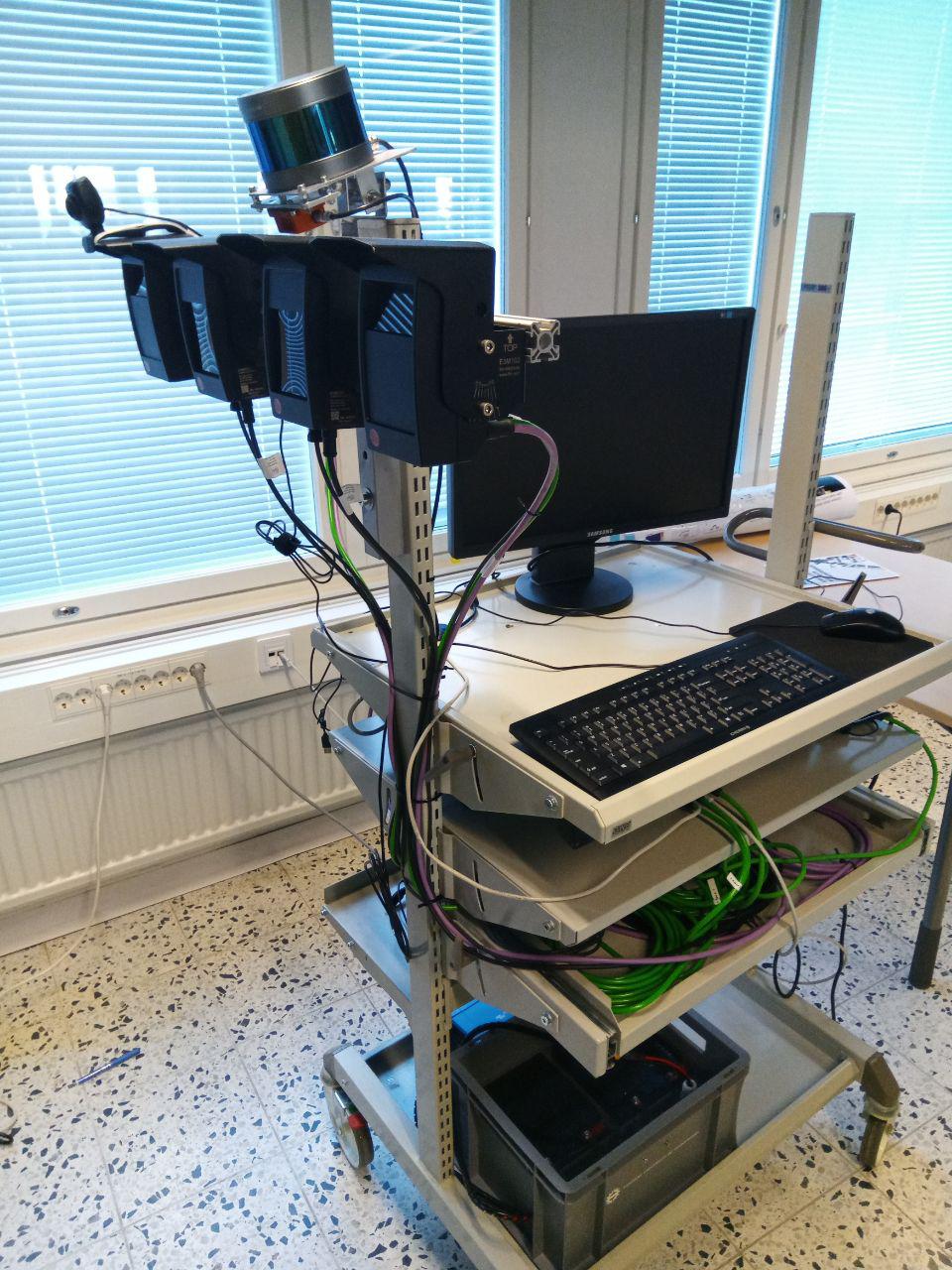}}
\caption{Data collection set-up. The IMU sensor is the orange box positioned under the LIDAR}
\label{fig_3}
\end{figure}

The data was collected using an industrial trolley with silent wheels shown in the Figure~\ref{fig_3}. The trolley has no motors and it was pushed across the corridors and offices. The machine was equipped with an Inertial Measurement Units (IMU sensor). The IMU sensor used is an XSENS MTi-300. The sensor data collected includes accelerometer data, gyroscope data (angular rate) and internally estimated orientation. Specifically:
\begin{itemize}
    \item[--] \textit{Orientation}: 4 attitude quaternion channels, 3 for vector part and one for scalar part;
    \item[--] \textit{Angular rate}: 3 channels, corresponding to the 3 orthogonal IMU coordinate axes X, Y, and Z;
    \item[--] \textit{Acceleration}: 3 channels, specific force corresponding to 3 orthogonal IMU coordinate axes X, Y, and Z.
\end{itemize}

While angular velocity and linear acceleration are given in the IMU body coordinates X, Y, and Z, the orientation is presented as a quaternion, a mathematical notation used to represent orientations and rotations in a 3D space~\cite{kuipers1999quaternions}.
The setup was driven through a total of 9 different surface types: hard tiles with large space, hard tiles, soft tiles, fine concrete, concrete, soft polyvinyl chloride (PVC), tiles, wood, and carpet.

Each data point includes the measures described above of orientation, velocity and acceleration, resulting in a feature vector of length 10 for each point. In total almost a million data points were collected.

\subsection{Data preparation}
\label{subsec:datapreparation}
In order to preserve the characteristics of the time series corresponding to different surface types, we grouped a fixed window of consecutive data point in order to obtain segments of length 128 points. After this process, our dataset includes 7626 segments.

The sample distribution by class is shown in Figure~\ref{fig_2}. From this, it can be seen that some of the classes are under-represented. In order to avoid that during the testing of the models some of the class were totally absent from the test set, we made sure during the cross validation process to have all classes represented in both the train and the test set. 

Moreover, we divided the data into \textit{groups} to be used for the cross validation. A group is a collection of subsequent 128-sample long blocks and in total we have 80 groups. We noticed, that a random train-test split would lead an overly optimistic accuracy estimate as neighboring samples would be in train and test sets. On the other hand, a systematic split by time (\textit{e.g.,} first third of each sequence for training, second for validation and third for testing) would probably give pessimistic and misleading accuracy estimates as the state (\textit{e.g.,} direction) could have changed radically. Thus, our compromise is to group adjacent segments together, and split the data such that each group belongs fully into one of the three folds, which makes the classification process fair and less biased.


\subsection{Setting up a machine learning competition}
\label{subsec:datacompetition}

The data was used in two public machine learning competition, both hosted on the Kaggle platform. The platform is free of charge for educational use, and these \textit{InClass} competitions have shown their importance for exposing machine learning students to practical machine learning challenges \cite{Gharib2018}. However, setting up the competition requires careful design as any coursework in order to encourage exploration and learning. 

The platform requires the data to be split in train, validation, and test folds. The validation and test set are used respectively to score the submission of the participants in the public and the private leaderboards. As the name suggests, the public leaderboard is visible to everyone for the whole duration of the competition and can be used for assessing the performance of one's own model. The private leaderboard is only visible once the competition ends and it is the basis for deciding the final scoring of the participant's model.  

It is not uncommon for the results to vary widely between the public and the private leaderboards, specifically if the data in the validation set is significantly different from the one in the test set. From an educational point of view, this can cause disappointment and discouragement for the young students participating in the competition, undermining the purpose of the course.
To avoid this, we split the data training a Gaussian Mixture Model (GMM)~\cite{spall1992feasible} for each of the set and measuring the symmetric Kullback-Leiber (KL) divergence~\cite{virtanen2007probabilistic}~\cite{heittola2014method}, in order for it to be minimized between the validation and the test set.
Specifically, we created 3 sets (train, validation and test) by random shuffling and dividing the segments, such that 50\% of the data was used for the training set, and 25\% each in the validation and the test sets.
We trained a GMM for each set and we measured the distance between them using the symmetric KL divergence. We did this iteratively for 1000 different splits, and chose the one with the minimum KL divergence between the validation and test set.

The orientation is estimated internally by the IMU by fusing accelerometer data, gyro data and optionally magnetometer data. In this process the gyro data is integrated over time, and furthermore, the integration is non-commutative~\cite{kuipers1999quaternions}. This means that segments can be identified with potentially high confidence by linking an end of one segment quaternion to a start of another segment quaternion. This is a potential source of leak for competitions where continuous data is scrambled. Furthermore, in practical applications the orientation is independent of surface type and orientation-based features can lead to a model that over-fits the training data. On the other hand, quaternions can be used to rotate the inertial measurements to locally level frame and this leads to improved sensitivity in surface detection. This leak was in fact recognized by some of the competitors, resulting in an unrealistically high accuracy, as we will discuss in the experimental section.

\section{Methods}
\label{sec:methods}
We study three alternative methods for  classifying the floor surface type. 
The first approach is a traditional combination of manual feature engineering coupled with a state-of-the-art classifier, with manually engineered features specifically tailored for this task. The second and third approach use a convolutional neural network, whose convolutional pipeline learns the feature engineering from the data. More specifically, the second model is a Fully Convolutional Network (FCN), while the third is a Residual Network (ResNet). These last two approaches are adopted from \cite{fawaz2019deep}, where the authors found their performance superior to a number of other tested methods. Specifically, \textit{Fawaz et al.} showed in their work that the FCN and the ResNet were the top performing methods for multivariate time series classification. This results were obtained testing 9 different classifiers on 12 multivariate time series datasets.

Besides studying each method singularly, we also study the fusion of the three methods by combining their predictions together in various ways. This method is a derivation of the winning model from our first, university level course, Kaggle competition. The winning team in fact used a combination of XGBoost and 1-dimensional convolutional networks to classify the surface type. 

All the tests in our study have been conducted using only six of the ten channels available in the dataset. In fact, we excluded the orientation channel because of the scarce correlation that it has to the surface type, as we discovered from the results and discussion of the two public competition. We still decided to include it in the dataset as it can be useful for future works.

\subsection{Manually engineered features}
The first method used is based on a XGBoost model~\cite{chen2016xgboost}, a state-of-the-art, optimized, implementation of the Gradient Boosting algorithm. It allows faster computation and parallelization compared to normal boosting algorithm. For this reason it can yield better performance compared to the latter, and can be more easily scaled for the use with high dimensional data.
In our case, we trained a tree based XGBoost model, with 1000 estimators and a learning rate of $10^{-2}$. 

This model uses the following human-designed basic features: We compute the \textit{mean}, the \textit{standard deviation}, and the \textit{fast fourier transform} (FFT) of each of the six measurement channels, and used these three features for the training of the model.
Specifically the FFT was included for its ability to simplify complex repetitive signals, highlighting their key components.



\subsection{Fully Convolutional Neural Network}
The second method used, is a fully convolutional neural network (FCN)~\cite{wang2017time}. This convolutional network does not present any pooling layer (hence the name), therefore the dimension of the time series remains the same through all the convolution. Moreover, after the convolutions, the features are passed to a global average pooling (GAP) layer~\cite{lin2013network} instead of the more traditional fully connected layer. The GAP layer allows the features maps of the convolutional layers to be recognised as category confidence map. Moreover, it reduces the number of parameters to train in the network, making it more lightweight, and reducing the risk of overfitting, when compared to the fully convolutional layer.

The FCN used in this work, adopted from \cite{fawaz2019deep}, consists of 3 convolutional blocks, each composed by a 1-dimensional convolution followed by a batch normalization layer~\cite{ioffe2015batch} and a rectified linear unit (ReLU)~\cite{nair2010rectified} activation function. The output of the last convolutional block are fed to the GAP layer, to which a traditional softmax is fully connected for the time series classification.

\subsection{Residual Network}
The last method used is a residual network (ResNet)~\cite{wang2017time}, composed by 11 layers of which 9 are convolutional. The main difference with the FCN described before is the presence of shortcut connection between the convolutional layers. This connections allow the network to learn the residual~\cite{he2016deep}. This allow the network to be more easily training, as the gradient flows directly through the connections. Moreover, the residual connection allow to reduce the vanishing gradient effect.

In this work we used the ResNet shown in \cite{fawaz2019deep}. It consists of 3 residual blocks, each composed of three 1-dimensional convolutional layers, and their output is added to input of the residual block. The last residual block, as for the FCN, is followed by a GAP layer and a softmax.

\section{Evaluation}
\label{sec:evaluation}

In this section we study the accuracy of the proposed methods using the indoor surface dataset.

\subsection{Accuracy metrics}
The accuracy of the methods is evaluated using two accuracy metrics. The first one is the classical accuracy which evaluates the fraction of corrected prediction over the total number of samples.

The second accuracy metric used is the area under the receiver operating characteristic curve (AUC
The receiver operating characteristic (ROC) curve, is used to illustrate the performance of a classifier (usually binary), plotting the value of true positive rate (TPR) against the false positive rate (FPR), for various discrimination threshold values. 
Once a ROC curve is found, the area under the curve (AUC) can be calculated and used as an accuracy metric. 

For our specific data, as the it can belong to 9 different classes, it was necessary to average the ROC curve for all the classes. This was done through \textit{macro} averaging the ROC for each class, giving the same weight to each of them. The AUC was then calculated from the macro-averaged ROC curve as for equation 

\subsection{Cross validation}
To properly evaluate the performance of the methods used, we used an iterated random cross validation based on groups of samples. As per Section~\ref{subsec:datapreparation}, we divided the data into 80 groups, each composed of subsequent samples belonging to the same class, and then randomly split the dataset into train and test set according to the groups. 
The random cross validation was iterated 5 times, and the data was split such that 70\% of the groups was present in the training set and 30\% in the test set. We made sure that each class was always present in both the train and the test set.
The accuracy metrics were calculated for each fold, after which their means and standard deviation were taken into account.

We choose to use this cross validation as we noticed that adjacent samples were similar. Using a standard cross validation, there was the risk that very similar samples were divided between the train and the test set, making the classification task biased. Using an iterated cross validation over groups of samples, we reduced this risk as similar segments are always either in the train or in the test set.


\subsection{Results}

\begin{table}[]
\begin{center}
\caption{Evaluation score for the models used.}
\begin{tabular}{l||ll}
                                          & Accuracy score & AUC score \\ \hline \hline
XGBoost          & $59.54\% \pm 5.57\%$ & $89.59\% \pm 1.51\%$ \\ 
FCN                  & $62.69\% \pm 6.74\%$ & $90.45\% \pm 2.20\%$ \\
ResNet                   & $64.95\% \pm 3.39\%$ & $\bf 92.33\% \pm 1.16\%$ \\
XGB + FCN                   & $64.87\% \pm 6.14\%$ & $91.26\% \pm 1.77\%$ \\
XGB + ResNet                   & $65.76\% \pm 3.97\%$ & $91.40\% \pm 1.46\%$ \\
FCN + ResNet                   & $67.57\% \pm 4.34\%$ & $92.30\% \pm 1.62\%$ \\
XGB + FCN + ResNet                   & $\bf 68.21\% \pm 5.12\%$ & $ 91.98\% \pm 1.65\%$            
\end{tabular}

\label{table_1}
\end{center}
\end{table}

We implemented the models presented in Section~\ref{sec:methods} in Python, using Keras API~\cite{chollet2015keras} for training the deep learning models, and Scikit-Learn~\cite{scikit-learn} for the XGBoost method. 
We trained each model individually and then combining them using a voting system, which gives each model used the same weight. We consider therefore all possible combination of the three models, both in pairs and considering all 3 of them together.

The models were trained on 70\% of the data and tested on the remaining 30\%, for 5 times using an iterated random cross validation. The results from each fold was then averaged, and the final score for each model can be seen in Table~\ref{table_1}.
The deep learning models performed well, all of them achieving an accuracy of over 60\% and an AUC of over 90\%. 
The best performing method is the combination of all the three models, with an accuracy of 68.21\% and an AUC of 91.98\%. Shortly behind is the combination of the two deep learning models, FCN and ResNet, with an accuracy of 67.57\% and AUC of 92.30\%.
The individual neural networks coupled with the XGBoost achieved an accuracy of 64.87\% for the FCN and XGBoost, and 65.76\% for ResNet and XGBoost, with an AUC of 91.26\% and 91.40\% respectively.

\begin{figure}[t!]
\centerline{\includegraphics[scale=0.64, trim={0 0 0 1cm},clip]{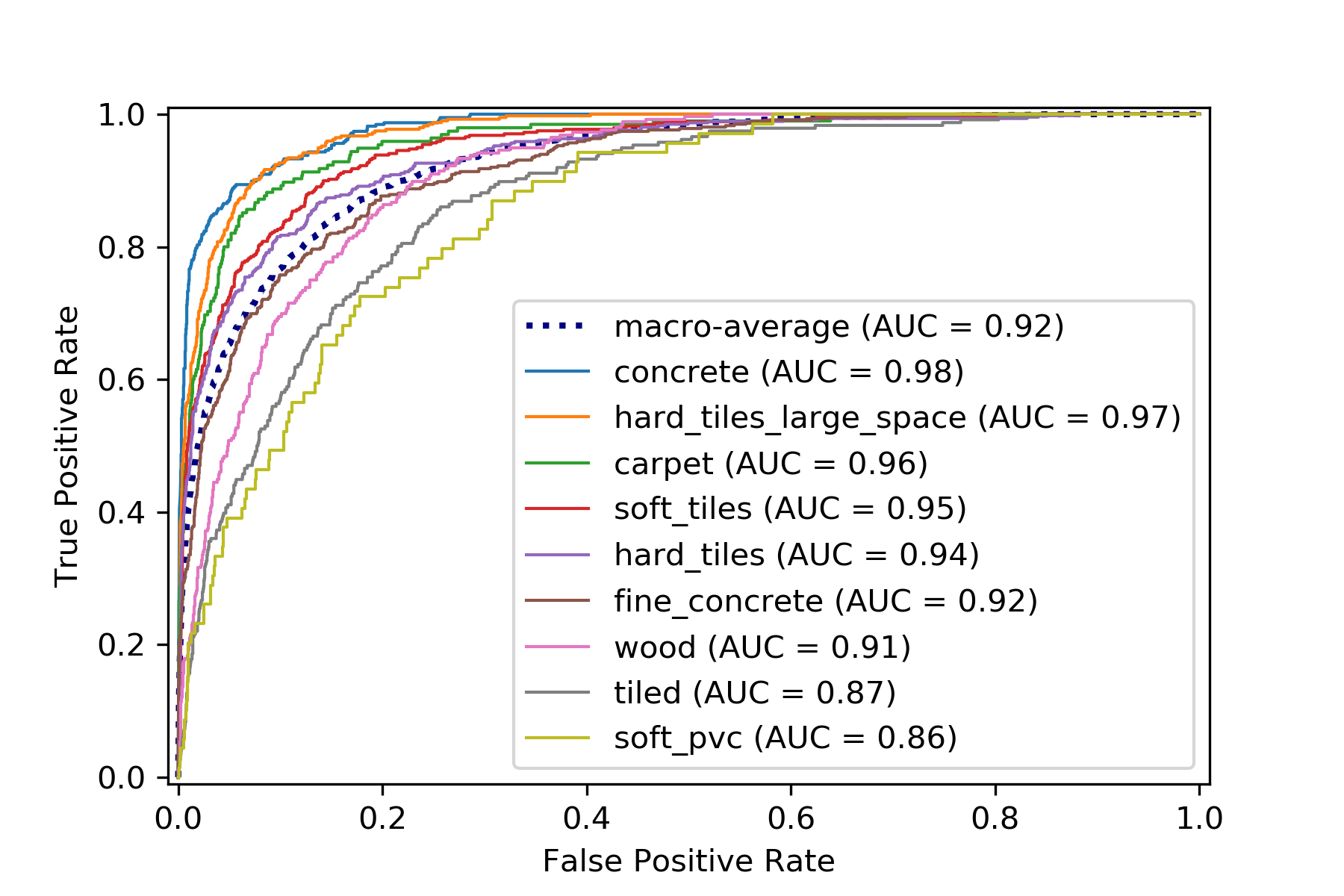}}
\caption{ROC curve with AUC values for each class and for the macro averaging.}
\label{fig_4}
\end{figure}

Moreover, as it can be seen from Table~\ref{table_1}, combining models together yielded a higher accuracy and AUC in all case. It can be also noted that the ResNet alone achieved a remarkably high accuracy when compared with the other methods (64.95\%), and the highest AUC (92.33\%), showing that this state-of-the-art neural network architecture is capable of learning faster and better features from the time series. Figure~\ref{fig_4} shows the ROC curve for each class and its macro averaging for the ResNet model.

\subsection{Competition Results}
Following closely the results and analysis performed by the participant of the competitions, we noticed that the top performing team used methods based on 1-dimensional CNN. Their results are much higher than the one we are presenting in this paper, specifically if we look at the \textit{Kaggle} \textit{CareerCon} competition's results. Here, the top scores are comprised between 90\% and 99\%, we the winning participant scoring 100\% accuracy. Looking closely at the results we noticed that this was mainly due to the use of the Orientation channels of the IMU sensor: the orientation is less linked to the surface type itself, and more on how the trolley is moving. For this reason, many teams were able to spot similarities between the data in the training set and in the test set, and therefore understand with good approximation to which class the samples in the test set belonged.  

\section{Conclusions}
\label{sec:conclusion}
In this work we showed the collection of an IMU sensor data recorded through a wheeled robot driven indoor on different surface types. From these recordings, we built a dataset which, to the best of our knowledge, is unique in its scope and in its dimension.

The dataset contains data related to the orientation, angular velocity and linear acceleration of the robot moving on 9 different surface types. All over we collected roughly one million measurements which were then combined into 7626 labeled segments of 128 data points each.

We tested different state-of-the-art model for time series classification on the dataset, and we proposed a baseline method for our data based on an ensemble of state-of-the-art classifier trained on manually engineered features and two different deep learning models based on convolutional network. The baseline model scored an mean accuracy over 5-folds iterated random cross validation of 28.21\% and an AUC of 91.98\%.

The data was already used in two public competition, hosted on Kaggle platform: the first related to an advance machine learning university level course, and the second for a worldwide \textit{CareerCon} event organized by Kaggle itself. The competitions have reached a total of over 1500 participants combined.

\section{Acknowledgements}
The authors would like to thank Kaggle.com as the platform for organizing the two public competitions, and Dr. Walter Reade and Mr. Sohier Dane for insightful discussions during their preparation. The authors would also like to thank CSC - The IT Center
for Science for the use of their computational resources. The
work was partially funded by the Academy of Finland project
309903 CoefNet and Business Finland project 408/31/2018
MIDAS.

\bibliographystyle{unsrt}  
\bibliography{references}  

\begin{thebibliography}{10}

\bibitem{ojeda2006terrain}
Lauro Ojeda, Johann Borenstein, Gary Witus, and Robert Karlsen.
\newblock Terrain characterization and classification with a mobile robot.
\newblock {\em Journal of Field Robotics}, 23(2):103--122, 2006.

\bibitem{oliveira2017speed}
Felipe~G Oliveira, Elerson~RS Santos, Armando~Alves Neto, Mario~FM Campos, and
  Douglas~G Macharet.
\newblock Speed-invariant terrain roughness classification and control based on
  inertial sensors.
\newblock In {\em 2017 Latin American Robotics Symposium (LARS) and 2017
  Brazilian Symposium on Robotics (SBR)}, pages 1--6. IEEE, 2017.

\bibitem{kertesz2016rigidity}
Csaba Kert{\'e}sz.
\newblock Rigidity-based surface recognition for a domestic legged robot.
\newblock {\em IEEE Robotics and Automation Letters}, 1(1):309--315, 2016.

\bibitem{walas2015terrain}
Krzysztof Walas.
\newblock Terrain classification and negotiation with a walking robot.
\newblock {\em Journal of Intelligent \& Robotic Systems}, 78(3-4):401--423,
  2015.

\bibitem{brooks2012self}
Christopher~A Brooks and Karl Iagnemma.
\newblock Self-supervised terrain classification for planetary surface
  exploration rovers.
\newblock {\em Journal of Field Robotics}, 29(3):445--468, 2012.

\bibitem{valada2017deep}
Abhinav Valada and Wolfram Burgard.
\newblock Deep spatiotemporal models for robust proprioceptive terrain
  classification.
\newblock {\em The International Journal of Robotics Research},
  36(13-14):1521--1539, 2017.

\bibitem{kuipers1999quaternions}
Jack~B Kuipers et~al.
\newblock {\em Quaternions and rotation sequences}, volume~66.
\newblock Princeton university press Princeton, 1999.

\bibitem{Gharib2018}
Shayan Gharib, Honain Derrar, Daisuke Niizumi, Tuukka Senttula, Janne Tommola,
  Toni Heittola, Tuomas Virtanen, and Heikki Huttunen.
\newblock Acoustic scene classification: A competition review".
\newblock In {\em Proc. IEEE Symp. Machine Learn. for Signal Process., MLSP
  2018}, Sept. 2018.

\bibitem{spall1992feasible}
James~C Spall and John~L Maryak.
\newblock A feasible bayesian estimator of quantiles for projectile accuracy
  from non-iid data.
\newblock {\em Journal of the American Statistical Association},
  87(419):676--681, 1992.

\bibitem{virtanen2007probabilistic}
Tuomas Virtanen and Marko Hel{\'e}n.
\newblock Probabilistic model based similarity measures for audio
  query-by-example.
\newblock In {\em 2007 IEEE Workshop on Applications of Signal Processing to
  Audio and Acoustics}, pages 82--85. IEEE, 2007.

\bibitem{heittola2014method}
Toni Heittola, Annamaria Mesaros, Dani Korpi, Antti Eronen, and Tuomas
  Virtanen.
\newblock Method for creating location-specific audio textures.
\newblock {\em EURASIP Journal on Audio, Speech, and Music Processing},
  2014(1):9, 2014.

\bibitem{fawaz2019deep}
Hassan~Ismail Fawaz, Germain Forestier, Jonathan Weber, Lhassane Idoumghar, and
  Pierre-Alain Muller.
\newblock Deep learning for time series classification: a review.
\newblock {\em Data Mining and Knowledge Discovery}, pages 1--47, 2019.

\bibitem{chen2016xgboost}
Tianqi Chen and Carlos Guestrin.
\newblock Xgboost: A scalable tree boosting system.
\newblock In {\em Proceedings of the 22nd acm sigkdd international conference
  on knowledge discovery and data mining}, pages 785--794. ACM, 2016.

\bibitem{wang2017time}
Zhiguang Wang, Weizhong Yan, and Tim Oates.
\newblock Time series classification from scratch with deep neural networks: A
  strong baseline.
\newblock In {\em 2017 International joint conference on neural networks
  (IJCNN)}, pages 1578--1585. IEEE, 2017.

\bibitem{lin2013network}
Min Lin, Qiang Chen, and Shuicheng Yan.
\newblock Network in network.
\newblock {\em arXiv preprint arXiv:1312.4400}, 2013.

\bibitem{ioffe2015batch}
Sergey Ioffe and Christian Szegedy.
\newblock Batch normalization: Accelerating deep network training by reducing
  internal covariate shift.
\newblock {\em arXiv preprint arXiv:1502.03167}, 2015.

\bibitem{nair2010rectified}
Vinod Nair and Geoffrey~E Hinton.
\newblock Rectified linear units improve restricted boltzmann machines.
\newblock In {\em Proceedings of the 27th international conference on machine
  learning (ICML-10)}, pages 807--814, 2010.

\bibitem{he2016deep}
Kaiming He, Xiangyu Zhang, Shaoqing Ren, and Jian Sun.
\newblock Deep residual learning for image recognition.
\newblock In {\em Proceedings of the IEEE conference on computer vision and
  pattern recognition}, pages 770--778, 2016.

\bibitem{chollet2015keras}
Fran\c{c}ois Chollet et~al.
\newblock Keras.
\newblock \url{https://keras.io}, 2015.

\bibitem{scikit-learn}
F.~Pedregosa, G.~Varoquaux, A.~Gramfort, V.~Michel, B.~Thirion, O.~Grisel,
  M.~Blondel, P.~Prettenhofer, R.~Weiss, V.~Dubourg, J.~Vanderplas, A.~Passos,
  D.~Cournapeau, M.~Brucher, M.~Perrot, and E.~Duchesnay.
\newblock Scikit-learn: Machine learning in {P}ython.
\newblock {\em Journal of Machine Learning Research}, 12:2825--2830, 2011.

\end{thebibliography}

\end{document}